\documentclass{tlp}

\usepackage{amssymb}

\usepackage{graphicx}
\usepackage{cleveref}

\usepackage{url}

\newcommand{\comment}[1]{}

\begin{document}

\title{A Review of Literature on Parallel Constraint Solving}


\author[Ian P. Gent et al.]
{IAN P. GENT\\
School of Computer Science, University of St Andrews, St Andrews KY16 9SX, UK\\
\email{ian.gent@st-andrews.ac.uk}
\and CIARAN MCCREESH\\
School of Computing Science, University of Glasgow, Glasgow G12 8RZ, UK\\
\email{ciaran.mccreesh@glasgow.ac.uk}
\and IAN MIGUEL\\
School of Computer Science, University of St Andrews, St Andrews KY16 9SX, UK\\
\email{ijm@st-andrews.ac.uk}
\and NEIL C.A.\ MOORE\\
Adobe Systems Incorporated, Edinburgh, UK\\
\email{nemoore@adobe.com}
\and PETER NIGHTINGALE\\
School of Computer Science, University of St Andrews, St Andrews KY16 9SX, UK\\
\email{pwn1@st-andrews.ac.uk}
\and PATRICK PROSSER\\
School of Computing Science, University of Glasgow, Glasgow G12 8RZ, UK\\
\email{Patrick.Prosser@glasgow.ac.uk}
\and CHRIS UNSWORTH\\
13 Grasmere Drive, York, YO10 3RY, UK\\
\email{chris.unsworth79@gmail.com}}

\maketitle

\begin{abstract}
As multicore computing is now standard, it seems irresponsible for constraints researchers to ignore the implications of it. Researchers need to address a number of issues to exploit parallelism, such as: investigating which constraint algorithms are amenable to parallelisation; whether to use shared memory or distributed computation; whether to use static or dynamic decomposition; and how to best exploit portfolios and cooperating search. 
We review the literature, and see that we can sometimes do quite well, some of the time, on some instances, but we are far from a general solution. Yet there seems to be little overall guidance that can be given on how best to exploit multicore computers to speed up constraint solving. We hope at least that this survey will provide useful pointers to future researchers wishing to correct this situation.

Under consideration in Theory and Practice of Logic Programming (TPLP).
\end{abstract}

\newpage
\section{Introduction}

How can constraint solvers best exploit parallel processing when all workstations, laptops, tablets and even phones are multicore computers?  To address this question we review the literature on exploiting parallel processing in constraint solving.
We start by looking at
the justification for the multicore architecture, the direction it is most likely to take and limiting factors on performance. We then review recent literature on parallel constraint programming and SAT solving. We have organized the survey into four categories, as follows.

\begin{itemize}
\item Parallel consistency and propagation (\Cref{sec:parprop}), where constraint propagation algorithms are parallelized.
\item Multi-agent search (\Cref{sec:multiagent}), where multiple agents attempt to solve the problem in parallel while sharing useful information.
\item Parallelizing the search process (\Cref{sec:parsearch}), in which the search process is split among multiple workers in some way.
\item Portfolios (\Cref{sec:portfolios}), where a set of diverse solvers are selected to run in parallel until one of them solves the problem. 
\end{itemize}


Our survey is focused on approaches to parallelism developed in the constraint programming and SAT communities.  
A number of surveys of related areas have previously appeared, whose contributions we gratefully acknowledge and cite below where they overlap with our concerns. 
These include surveys of parallel solving in SAT \cite{sat-survey-12,sat-survey-icacsis11,sat-survey-singer06}, Distributed Constraint Satisfaction \cite{Yokoo2000,DBLP:reference/fai/Faltings06}, algorithm selection and portfolios \cite{kotthoff2014algorithm}, Concurrent Constraint Programming \cite{DBLP:reference/fai/FruhwirthMS06}, and a proposal of seven challenges for future research in parallel SAT \cite{hamadi2013seven}.

Other surveys address closely related problems amenable to parallelisation, but fall outside the scope of this paper's focus on constraint solving. For example, Mixed integer linear programming (MILP) is a powerful technique from the operations research community for solving
discrete optimization problems: Ralphs et al~\citeyear{mip-survey-17} have written an excellent survey of parallel MILP solving. 

\section{The Hardware: Multicore, GPU and Amdahl's Law}

Written in 2006, Intel's White Paper \cite{teraScale} starts by saying ``... two cores are here now, and quad cores are right 
around the corner''. Now, 16 and 32 core machines are commonly available.
But why go multi-core? In the past performance improvements could be taken for granted as clock speeds increased (from 5 MHz in 1978 to more 
than 4 GHz in 2018), component size decreased, and chip density increased. Three reasons are given for the shift to multi-core. First, 
although component size continues to fall, power-thermal issues limit performance, so we can no longer simply increase clock 
speeds. Secondly, power consumption: individual cores can be tuned for different usages (e.g.\ dedicating hardware resources to 
specific functions), and when not in use cores can be powered down. And thirdly, rapid design cycles: hardware designs can be 
reused across generations. 

What are the major challenges? Intel put top of their list ``programmability'', that the platform must 
address new and existing programming models. And then ``adaptability'', such that the platform can be dynamically reconfigured 
to conserve power. Of course ``reliability'', ``trust'', and ``scalability'' are also important, as we increase cores we cannot compromise the correctness of the hardware.

Intel considers development of multi-core software to be amongst the greatest challenges for tera-scale computing, specifically
with regard to ensuring that ``there are compelling applications and workloads that exploit the massive compute density'' and
that ``multiprocessing adds a time dimension that is extremely difficult for software developers to cope with''. They give
a further justification for the multi-core architecture: ``... why tomorrow's applications need so many threads. The answer is that
those advanced, intelligent applications require supercomputing capabilities, and the accompanying parallelism that allows those 
applications to proceed in real-time. ... it requires an equally massive shift in hardware and software.''

Intel's tera-scale computing vision is to aim for hundreds of cores on a chip, giving the capability of performing trillions
of calculations per second on trillions of bytes of data with a stated goal ``... a 10$\times$ improvement in performance per watt 
over the next 10 years.'' \cite{teraScale}. How close are we to that goal? In 2006 two core machines existed. At the time of writing in 2017, Intel are selling server chips with 24 physical cores (48 threads with hyperthreading) running at 2.40 GHz. Their Teraflops Research Chip (Polaris) contains 80 cores.

But there is a shadow cast over this optimism: Amdahl's law.
Amdahl's law predicts the maximum speedup that can be expected from a system as we increase the number of
processors. The law assumes that a program is composed of a parallel part $P$ and a sequential
part $S$, such that $P+S=1$. The expected speed up is then $1/(S+P/N)$, where $N$ is the number
of processors. As $N$ tends to infinity Amdahl's law predicts that maximum speedup will be $1/S$, 
as the original $P/N$ term tends to zero. As an example if we had $P = 0.99$, so 99\% of our problem can parallelised, 64 processors
would run our program 39 times faster.
For 128 processors the speedup is 56 times, for 1024 processors it is 91. As the number of processors continues to increase the speedup tends
to 100. If $P = 0.9$ the law predicts a maximum speed up of 10, and if half only our program
can be parallelized, $P = 0.5$ and maximum speed up is 2, regardless of the number of processors available.
It was this argument, in the late 1960's, that encouraged hardware development away from multi-processor
and towards faster processors.

In the late 1980's Gustafson \citeyear{gustafson} argued that Amdahl's law is overly pessimistic, as it assumes that as we increase
the available parallel processors we continue to keep the workload fixed and hope for reduced runtime. That is,
it is a ``fixed-size speedup" model and assumes $N$ and $P$ are independent; multi-processing is only used
to improve response time. Gustafson assumes that problem size also
scales with the number of processors, i.e. as we get more processors we increase the problem size and that run time,
not problem size, is a constant. Gustafson observed that the parallel or vector parts of a program scales with
problem size and the serial part does not (it diminishes proportionally). Consequently as we get more
processors the workload grows and $P$ increases resulting in an increase in speedup. This is the
``fixed-time speedup" model and an example is weather forecasting, where we use multi-processors to
increase the quality of our results (the weather prediction) in a fixed amount of time (before the evening news).  Perhaps this model is more appropriate for parallel constraint solving where we are always striving to solve larger and harder instances.  

The constraints community has a long history of engaging with the challenges involved in concurrent and parallel programming. Concurrent Constraint Logic Programming was developed in the 1980s \cite{DBLP:conf/iclp/Maher87,DBLP:conf/fsttcs/FurukawaU88} followed by, outside the context of logic programming, Concurrent Constraint Programming
\cite{Saraswat:1989:CCP:96709.96733,DBLP:conf/ijcai/HenzSW93,DBLP:reference/fai/FruhwirthMS06}. This led to incorporation of constraint reasoning into multi-paradigm languages such as Mozart \cite{van1803mozart}.  However, ongoing developments in efficient constraint solving have meant that constraint techniques now seem less suitable for integration into the heart of languages like Mozart. As a result, constraints are no longer available in the first release of Mozart 2: It is planned to retain constraint solving in Mozart via linkage to the constraints library Gecode \cite{gecode}.

One current area of great interest is solving on GPUs. Almost all modern desktops and laptops provide a powerful GPU, and there are several popular methods of utilising GPUs, including CUDA\footnote{\url{http://www.nvidia.com/object/cuda_home_new.html}} and OpenCL\footnote{\url{https://www.khronos.org/opencl/}}. Using GPUs has led to orders of magnitude improvement on many important problems, including \(k\) nearest neighbour \cite{10.1109/CVPRW.2008.4563100}, MaxSAT \cite{Munawar:2009:HGA:1666141.1666143}, SAT \cite{Manolios2006Implementing,dalpalu2015gpu} and Constraint-Based Local Search \cite{arbelaez2014gpu}. One common thread in these papers is that applying a GPU provides the greatest improvements on problems which can be solved by massively parallel simple calculations. GPUs are not a silver bullet, and direct ports of existing algorithms to a GPU architecture often perform poorly.

\section{Parallel Consistency and Propagation}\label{sec:parprop}

In 1990 	 \cite{kasif90} showed that the problem of 
establishing arc-consistency (AC) is P-complete i.e.~the 
problem is not inherently parallelisable under the usual complexity assumptions. 
This is done by giving log-space reductions of AC to Horn-clause satisfaction and vice versa.
A major open problem in complexity theory is whether NC=P, where NC is ``Nick's Class", the class of problems that can be solved in polylog time using polynomially many processors. 
Kasif therefore showed that in the worst case we cannot establish arc-consistency 
exponentially faster with a polynomial number of processors, unless NC=P. This is no surprise, as we have to 
do a chain of deductions in arc-consistency, where each depends on (some subset of) the preceding ones.
We can read this result as being fatal to the enterprise of parallel consistency, 
but then, it is not fatal to solving constraint problems that 
they are NP-complete! So we have to take P-completeness into account rather than regard it as fatal.

While the general case of AC is P-complete, researchers have found special cases of problems that are in NC. While solving CSPs whose constraint graph is acyclic was known to be in P \cite{DBLP:journals/jacm/Freuder82}, Zhang and Mackworth showed that it was also in NC for constraints of arbitrary arity \cite{ZhangMackworth91}. Kasif and Delcher analysed a wider range of restrictions on the constraint graph \cite{DBLP:journals/ai/KasifD94}, while for arbitrary constraint graphs Kirousis gave a restriction on constraint relations being used which also leads to membership of NC \cite{KIROUSIS1993147}.

\subsection{Parallel Arc-Consistency}

There has been a steady stream of work on distributed consistency algorithms. One of the justifications of this is that the 
problem itself may be distributed geographically or due to organisational structures, as in  
 \cite{dcms92} and discussed further in \Cref{sec:dcsp}. The other 
justification is speed, which we discuss here. Here, 
the P-completeness of arc-consistency need not be fatal: it does not exclude the possibility of obtaining useful speedup from parallel processing.

For binary constraints, Kasif and Delcher showed that if a problem has $n$ variables with domain size $K$, then AC can be solved in $O(nK)$ time using $O(nK)$ processors, as long as the constraint graph contains $O(n^2)$ constraints
\cite{DBLP:journals/ai/KasifD94}.
Nguyen and Deville presented a distributed AC-4 algorithm DisAC-4 \cite{nguyen95,nguyen98}.
The algorithm is based on message passing. The variables are partitioned among the 
workers, and each worker essentially maintains the AC4 data structures for its
set of variables. When a worker deduces a domain deletion, this is broadcast to
all other workers. Each worker maintains a list of domain deletions to process
(some generated locally and others received from another worker). The worker
reaches a fixpoint itself before broadcasting any domain deletions, 
and waiting for new messages from other workers. The whole system reaches
a fixpoint when every worker has processed every domain deletion. 
It may be a difficult problem to partition the variables such that the work
is evenly distributed. The experimental results are mixed, with some experiments
showing close to linear speedup, while others show only 1.5 times speedup
with 8 processors.  A similar approach led to algorithms DisAC-3 and DisAC-6 based on their 
sequential counterparts AC-3 and AC-6 \cite{baudot1997analysis}.
Hamadi \citeyear{hamadi2002} presented an optimal distributed AC algorithm, DisAC-9, optimal with respect to message 
passing whilst outperforming the fastest centalized algorithms.

\subsection{Parallel Propagation of Non-Binary Constraints}

Ruiz-Andino, Araujo, S\'aenz and Ruz \citeyear{iclp98} presented a 
distributed propagation algorithm for $n$-ary functional constraints. 
These constraints are represented as \textit{indexicals}, 
where each variable in the scope of the constraint has a functional expression defining its domain. For example, given the constraint \(x_1=x_2+4\), the indexical for \(x_1\) is \(x_1 \in \{ \mathrm{min}(v_2)+4 \ldots \mathrm{max}(v_2)+4\}\).
The CSP is split into $n$ subsets such that each constraint appears in exactly one subset. If a variable is associated with 
constraints in more than one set then that variable is duplicated. Each subset is propagated sequentially by its own processor 
and any domain reductions of variables shared between processors is communicated between processors.
The experiments presented show the relative performance gains by increasing the number of cores they make available to their 
algorithm. First consistency is established, then a variable assignment is made and consistency is re-established. This is repeated until a solution 
is found or a variable domain is wiped-out. The performance of this technique is highly dependent on the quality of the 
distribution of the CSP, which is a difficult problem 
in itself. The conflicting optimisation criteria for quality of a constraint distribution are minimising the network traffic 
whilst maximising the distribution of the propagation frontier. It appears that this technique will not handle high arity 
constraints well due to increased communication cost.

Parallel propagation has been proposed for numerical problems, where the variable 
domains are infinite. Domains are represented as an interval using two floating-point 
numbers, and the objective of propagation is to narrow the intervals.
Although numerical constraint satisfaction is not the focus of this survey, we would like to mention one paper.
Granvilliers and Hains \citeyear{parallel-interval-prop-2000} proposed a parallel propagation algorithm
for non-linear constraints. This was evaluated on a Cray multiprocessor. The 
gain from using 64 processors (compared to 1 processor) varies from almost 
nothing to about 6 times, depending on the problem instance.

Rolf and Kuchcinski \citeyear{trics2010} parallelise both search and consistency (we discuss parallelisation of search in \Cref{sec:parsearch}). 
They take a different approach to parallelising consistency, by splitting the set of constraints to be propagated among threads, 
rather than parallelising the work of a single constraint. They begin with an example demonstrating that a simple search parallelism 
scheme is at the mercy of the location of the solution(s) in the search tree. If they are all going to be found by the first thread anyway, 
the others are just adding overhead. They introduce some terminology: \textit{parallel search}, a type of OR parallelism and therefore data parallelism; and \textit{parallel 
consistency}, a type of AND parallelism and therefore task parallelism. They claim that, for many models, solvers spend an order of magnitude more time 
enforcing consistency than they do searching, in which case data parallelism is less suitable. Another flaw is that data parallelism
 naturally puts more stress on the memory bus \cite{sunChen}.
In their approach to parallel consistency, they require synchronization of pruning, but do not share data during 
pruning to avoid upsetting the internal data structures of global constraints. Rather than fixing which threads deal with which 
constraints, at each node each consistency thread takes a set of constraints to propagate from the queue. When all constraints in the
queue have been processed, updates are actually committed. The process can stop early if one of the threads detects inconsistency. 
When combining both parallel search and consistency, each search thread gets an associated set of consistency threads. An alternative 
architecture is briefly discussed in which all threads take from a shared work pool, but the authors claim that scheduling uptake 
from this pool could be prohibitively complex.
Experiments are on Sudoku and \(n\)-Queens using up to 64 threads on 8 cores. The gains are modest. They identify three problems: 
inefficiencies in parallel consistency caused by not sharing data, the synchronization of pruning described, and third the memory bus.

Campeotto et al \citeyear{campeotto14} investigated parallel propagation using a GPU architecture and the NVIDIA CUDA programming model. Each constraint is assigned its own block of threads on the GPU, and some propagators are further parallelised by filtering each variable in a separate thread. Also, the constraints may be divided between the host CPU and the GPU to add another level of parallelism. 
The work focuses on efficient propagation of the inverse constraint and the positive table constraint.  Modest speedups are reported when using the GPU (compared to the host CPU alone), with the highest being 6.6 times. 


\subsection{Parallel Unit Propagation in SAT}

Conflict-Driven Clause Learning (CDCL) SAT solvers are the most successful class of SAT solvers for structured instances, and efficient unit propagation is central to their success. Dal Pal\`u et al~\citeyear{dalpalu2015gpu} proposed a parallel unit propagation algorithm implemented on a GPU architecture, demonstrating approximately one order of magnitude speed-up compared to the host CPU. However they used an exotic unit propagation algorithm without watched literals. They also implemented a version with watched literals but noted that the speedup compared to the sequential version was negligible. It is not clear whether their techniques could be applied to a modern SAT solver with watched literal unit propagation. 

Manthey proposed a method to parallelise unit propagation using a multi-core shared memory architecture~\citeyear{manthey11}. One thread drives the CDCL search and performs all operations except unit propagation sequentially. Other threads are only active when the solver is performing unit propagation. The set of clauses is partitioned among the threads. Each thread propagates its own clauses and records any implied literals in its own queue. Then a thread checks all other threads for new implied literals in a way that is lock- and wait-free, and copies them into its own queue for processing. He reported a speedup of 1.57 times using two threads, and found that the approach does not scale beyond two threads. The method has been slightly refined in a later work by the same author \cite{manthey11improved}.

\subsection{Parallel Update in Local Search}
\label{puils}

Local search methods such as  Constraint-Based Local Search (CBLS)~\cite{cblsbook} start with a complete assignment that may violate some of the constraints. At each step, a change is made to the assignment with the goal of converging on a satisfying assignment (optionally optimising some criteria). 
For parallel implementation of local search, Verhoeven and Aarts \citeyear{Verhoeven1995} introduced the notion of `multi-walk' and `single-walk'.  A multi-walk search uses parallelisation to explore multiple parts of the search space at the same time, with parallel independent or loosely interacting local search processes. 
A single-walk search is an inherently sequential search, but the calculation of neighbourhoods and/or update of search state may be performed in parallel.  
Most work on parallel local search has focused on multi-walk parallelisation, as described in Section~\ref{sec:parlocalsearch}.

An example of parallel state update in local search is GENET, a neural network local search method for CSPs \cite{wang1991solving,DBLP:conf/aaai/DavenportTWZ94}. The convergence cycle in GENET involves each node updating in parallel \cite{DBLP:conf/aaai/DavenportTWZ94}. GENET was designed to be implemented on VLSI hardware \cite{wang1992cascadable}. 
Negative results on parallelising update within the Adaptive Search method were reported in the 
Partitioned
Global Address Space model
\cite{Munera2014}.

Single-walk and multi-walk approaches can be combined: i.e.\ a multi-walk search can be implemented in which individual search is itself parallelised. The use of GPUs is attractive in this case, given the cost effectiveness per thread and also the fact that the thread architecture of GPUs matches well with the architecture of local search \cite{arbelaez2014gpu}.  CBLS has been implemented using a GPU for the number partitioning, magic squares, and Costas array problems, obtaining promising results  \cite{arbelaez2014gpu}.  Speedups of up to 17 times were obtained for the first two problem classes, with a much lesser speedup for the Costas array problem since the neighbourhoods were so small that a pure multi-walk approach was used.

Large neighbourhood search (LNS) is a powerful local search technique for constraint optimisation problems. Given a complete assignment that satisfies all constraints, conventional LNS attempts to improve its objective value by \textit{relaxing} (unassigning) a subset of the variables (called a neighbourhood) and searching for an improved assignment within that neighbourhood. Campeotto et al \cite{campeotto2014gpulns} parallelise LNS on a GPU architecture, firstly by exploring multiple neighbourhoods in parallel and secondly by parallelising the search within each neighbourhood. Promising results are presented where the GPU LNS algorithm is compared to a CPU implementation of the same, and compared to a conventional LNS implementation in a CP solver. 

\subsection{Parallel Singleton Arc-Consistency}

In \cite{Gharbi15} a master/worker architecture is proposed where the master performs a backtracking search
and workers compute a high level of consistency, one that is not normally considered economical. That is, the master 
performs a relatively shallow inferencing search (i.e. maintaining generalised arc-consistency) while workers perform deep inference (i.e. singleton arc-consistency), 
communicating with each other via a collection of shared stacks. The architecture might be thought of as the workers being deep thinkers capable of interrupting an unencumbered searcher. The empirical study gives inconclusive results, but does point the way to exploiting this architecture
with various levels of consistency, not just SAC.

\section{Multi-Agent Search}\label{sec:multiagent}

In multi-agent search we have one problem and a collection of cooperating problem solving agents that execute in parallel. The agents may be diverse, and in fact diversity is a desirable property. When multi-agent search is applied to conventional constraint satisfaction problems, each agent has a copy of the whole problem and is capable of solving the problem independently.  The agents work on their own copy of the problem and they collaborate in some way. 

Multi-agent search is also applied to the Distributed Constraint Satisfaction Problem (DCSP) where each agent sees only part of the problem, and therefore no single agent can solve the entire problem alone. We give an overview of DCSP solving techniques in \Cref{sec:dcsp}.

Assuming we have $n$ processors, a speedup of less than $n$ is sub-linear, equal to $n$ linear, greater than $n$ super-linear.
Probably the first report of super-linear speedup is due to Rao and Kumar \citeyear{raoKumar88} in parallel depth-first search on the 
15-puzzle. They argue that if all solutions are uniformly distributed about the state space then average speedup can be super-linear. 
The next body of work to report the phenomenon of super-linear speedup was multi-agent search.  One of the earliest examples of this is due to Clearwater, Huberman and Hogg \citeyear{chh91}. To demonstrate the power of cooperative problem solving they investigated the time to solve word puzzles, posed as constraint satisfaction problems, using a collection of agents. Each agent could solve the problem independently. Agents wrote \textit{hints} to a shared blackboard, and agents randomly read hints from the blackboard whilst solving the problem. As the number of agents increases, and the diversity amongst agents increases, a \textit{combinatorial implosion} occurs with a subsequent super-linear speedup in problem solving. They present as an explanation of this phenomena ``... the appearance of a lognormal distribution in the effectiveness of an individual agent's problem solving. The enhanced tail of this distribution guarantees the existence of some agents with superior performance.'' The idea of multi-agent search was further explored in the portfolio-based search proposed by Gomes and Selman~\citeyear{gomes2001algorithm}. A portfolio is a multi-agent search with no communication between the agents. Portfolios are surveyed in \Cref{sec:portfolios}.

\subsection{Multi-agent Search in SAT}\label{sub:multiagent-sat}

The SAT community has been quick to exploit multi-agent search. An excellent survey of parallel SAT solving by Martins, Manquinho and Lynce~\citeyear{sat-survey-12} identifies multi-agent search (named \textit{portfolios} in their paper) as one of two main approaches to parallelism in SAT, with search-space splitting being the other main approach (covered in \Cref{sub:parsearchlearning} below). H\"olldobler et al \citeyear{sat-survey-icacsis11} gave a short survey of complete parallel SAT solvers, including multi-agent approaches. An earlier survey is also available \cite{sat-survey-singer06} but is largely superseded by Martins et al \citeyear{sat-survey-12}.  
In this section we survey a small number of the most notable multi-agent SAT solvers, and refer the reader to the earlier surveys \cite{sat-survey-12,sat-survey-icacsis11,sat-survey-singer06} for more detail. 

We focus on Conflict-Driven Clause Learning (CDCL) SAT solvers because they have been the most successful on structured SAT instances in recent years. CDCL SAT solvers generate \textit{learned clauses} that are entailed by the original formula. Sharing these learned clauses is a key opportunity for multi-agent SAT solving.  

First we consider a line of work where a small number of agents communicate intensively through shared memory. 
ManySAT~\cite{manysat-09} exploits one of the main weaknesses of DPLL solvers, namely their sensitivity to parameter tuning, to create a set of diverse SAT solvers. Each SAT solver operates on the entire formula, and with an unrestricted search (i.e.\ the search space is not divided among the agents). In the original (1.0) version of ManySAT, the SAT solvers share learned clauses of length 8 or less. The length limit is intended to allow the most important learned clauses to be shared while avoiding the overhead of sharing all learned clauses. ManySAT has four agents, each with a hand-crafted set of parameter values. 

In the improved ManySAT 1.1~\cite{manysat11} each pair of agents has a dynamically adjusted length limit.  The limits are adjusted based on the rate that shared clauses are received and also on \textit{quality} (which is a measure of the relevance of shared clauses to the search process of the solver receiving them). ManySAT 1.5~\cite{manysat15} takes a somewhat different approach where two of the agents are \textit{masters} and the other two are \textit{slaves}. Each master directs the search of one slave in order to improve the quality of learned clauses transmitted from the slave to the master. 
In essence ManySAT is an invocation of Clearwater, Huberman and Hogg's cooperative problem solving strategy, but rather than share hints agents share nogoods, i.e.\ facts as to where solutions cannot exist. ManySAT has been successful in SAT competitions, suggesting that intensive clause sharing is an interesting strategy for shared memory systems. ManySAT 1.0 won the parallel track of SAT-Race 2008, while version 1.1 won the parallel track of the SAT 2009 competition. Version 1.5 came second in the parallel track of SAT-Race 2010.

Interleaving search with \textit{inprocessing}~\cite{jarvisalo2012inprocessing} has been shown to extend the reach of sequential CDCL solvers. Inprocessing simplifies the formula by applying a set of rules. An example is identifying two literals that take the same value in all solutions and replacing one literal with the other throughout. Inprocessing typically has a large set of configurations, so it can serve as another source of diversity. 

Plingeling is a multi-agent solver that exploits inprocessing. It builds on the highly efficient sequential solver Lingeling~\cite{plingeling2010,jarvisalo2012inprocessing} by running \(n\) versions of Lingeling in parallel with different random seeds, different configurations of inprocessing, and a different initial variable and value ordering. In the 2010 version of Plingeling~\cite{plingeling2010} only unit clauses (i.e.\ assignments of SAT variables) are shared between agents. Despite this very simple clause sharing scheme, Plingeling has been highly successful. Plingeling won the SAT-Race 2010 parallel track. Martins et al \citeyear{sat-survey-12} compared Plingeling with all three versions of ManySAT and a number of other (less successful) solvers. The solvers were allowed four parallel threads. Plingeling performed best overall despite having the worst speedup factor (compared to the sequential version of the same solver) of only 1.60. It seems the strength of the underlying solver is more important than parallelism in this case. 

Plingeling has continued to improve alongside the sequential solver Lingeling. In 2013 the authors introduced sharing of clauses up to length 40 \cite{plingeling2013}. Plingeling continues to do well in competitions: it came second in the parallel track of the SAT 2016 competition, after its sister solver Treengeling \cite{plingeling2013} (which is based on dividing the problem instance). 

All the above approaches rely on shared memory for fast communication between the agents. In contrast, Hyv\"arinen, Junttila and Niemel\"a \citeyear{CLSDSAT} investigated parallel SAT solving in distributed computing environments without shared memory. They report experiments with up to 96 parallel workers.  They proposed \textit{Clause Learning Simple Distributed SAT} (CL-SDSAT), where the technique is to run multiple independent randomized SAT solvers with no direct communication. Each worker is given a time limit. When a solver times out, it shares some of its learned clauses with the master. The shared clauses from workers are combined centrally, and whenever a new worker is started it is given the current set of shared clauses. 
Filtering the shared clauses is key to this approach. Hyv\"arinen et al \citeyear{CLSDSAT} propose that the workers should share their shortest clauses, and the central store should select clauses that have been learned independently by the largest number of workers. CL-SDSAT can be instantiated with any sequential SAT solver with minimal changes. 

An analysis of the runtime distribution of the randomized workers shows that the technique can perform well simply because some workers will have short runtimes. Also the clause sharing scheme is shown to reduce the expected runtime of workers. CL-SDSAT is not shown to achieve a linear or super-linear speedup, so it is probably most useful for very hard instances where an answer is required in a short time.  

Recent work by Balyo, Sanders and Sinz \citeyear{hordesat-2015} broadly follows the same approach of diversification and clause sharing. The system is designed for a cluster of computers each of which has multiple cores and shared memory, therefore communication takes advantage of shared memory when it is available. As in CL-SDSAT, short clauses are preferred for sharing. They report results of experiments on up to 2048 cores. Mean and median speedups are reported, and in some cases the mean speedup is super-linear however the median is sub-linear. 

Multi-agent search with sharing of learned clauses has been applied in a learning constraint solver by Ehlers and Stuckey \citeyear{DBLP:conf/cpaior/EhlersS16}. We discuss their work in \Cref{sub:parsearchlearning}. 

On a less positive note, a recent study has shown that resolution refutations (i.e.\ resolution proofs of unsatisfiability) produced by sequential SAT solvers are typically very deep and contain many \textit{bottlenecks} (depths where the proof contains a small number of clauses) that must be processed sequentially \cite{katsirelos2013resolution}. They conclude that it is impossible to produce such refutations with a high degree of parallelism, limiting the speedup of multi-agent search in SAT. The major assumption is that parallel solvers produce similar resolution refutations to sequential solvers. It is not clear how the findings apply to satisfiable instances, or indeed to parallel search (\Cref{sub:parsearchlearning}) where the solver produces resolution refutations in parallel for each fragment of the search space. 

Of the seven challenges posed by Hamadi and Wintersteiger \citeyear{hamadi2013seven}, the most relevant here is the challenge to improve estimates of the local quality of incoming (shared) clauses. Many solvers simply prefer short clauses or employ fixed limits on clause length or the value of a heuristic \cite{plingeling2013,hordesat-2015}. However there has been work on measuring the relevance of imported clauses \cite{Audemard2012}, and on managing imported clauses separately to reduce overheads and protect the importing thread from being swamped \cite{Audemard2012,audemard2014}. In our view the challenge has not been comprehensively addressed. 

To conclude, multi-agent search with shared clauses is a popular and successful approach to parallel SAT. How it compares to other approaches such as search splitting is the topic of on-going research.

\subsection{Distributed Constraint Satisfaction and Optimisation Problems}\label{sec:dcsp}

The Distributed Constraint Satisfaction Problem (DCSP) and its optimisation equivalent (DCOP) is an area of multi-agent systems that has been extensively researched over many years. DCSPs  and DCOPs have been solved using asynchronous backtracking techniques \cite{yokoo1992distributed} and also by distributed local search techniques \cite{DBLP:journals/ai/HirayamaY05,ZHANG200555}. Excellent surveys have has been written by Yokoo and Hirayama~\citeyear{Yokoo2000} and by Faltings \citeyear{DBLP:reference/fai/Faltings06}. 

There is a critical distinction between DCSPs and conventional 
CSPs - called `centralized' CSPs by Yokoo and Hirayama. 
In a DCSP, no agent holds the entire CSP: indeed each variable in the CSP is \textit{owned} by a given agent, and inter-agent constraints exist between variables held by different agents. Some of these constraints may not yet be known by an agent, and become known by message-passing. An example would be allocation of nurses to shifts in a hospital containing several departments: to a large extent each department can allocate its nurses independently, but there will be inter-departmental constraints which may invalidate a schedule proposed by one department.  Thus, although DCSP does address parallel constraint solving, the research motivation is different. Yokoo and Hirayama state:  
``Of course, even if the research motivations are different, the same algorithm might be useful for both. However, as far as authors know, existing parallel/distributed processing methods for solving CSPs are not suitable for distributed CSPs, since they usually require some global knowledge/control among agents.'' This conclusion has not remained universally true, 
since Hamadi's DisAC-9 algorithm has attracted interest from both points of view \cite{hamadi2002}. Nevertheless, most research on DCSP and DCOP has focussed on 
the case of a distributed problem rather than the distributed solution of a centralized problem. 
The latter is our focus here.

DCSP techniques were used to parallelise the search to a centralized CSP \cite{DBLP:journals/amc/SalidoB06}.  A graph partitioning algorithm was used to divide the original CSP into an appropriate number of subproblems, in this case 10, aiming to minimise the number of variables shared between subproblems.  The subproblems were then solved concurrently, with communication between agents to ensure consistency on the shared variables.

We do not attempt to survey the large amount of ongoing research into non-centralized DCSP and DCOPs. Research  has continued intensively with a particular focus on the optimisation variant DCOP, such as for example
\cite{DBLP:journals/jair/GrinshpounGZNM13,Wahbi2014,Zivan2015,Netzer2016,Sassi201744,FiorettoLYPS14,FiorettoLP0S15,Fioretto0P16}. 
Significant systems have been built for reasoning on DCSPs and DCOPS, including FRODO \cite{FRODO2}
and DisChoco \cite{Wahbi2014}.

\section{Parallelizing the Search Process} \label{sec:parsearch}

By ``parallelizing the search process'' we mean search parallelism at the granularity of nodes or search states, hence
each worker is close to being a standard sequential constraint process but they are collectively orchestrated to be part
of the same overall search process. We refer below to both local and complete search. For local search, parallelism
has been used to increase the number of starting points available in the local search \cite{cp2006}. For complete
backtracking search, the issues we will highlight include:

\begin{itemize}
\item how the search  space is divided between workers;
\item how workers communicate what portions of search they have completed and what new solutions and improvements to
their optimisation function they have found;
\item how state is shared (if appropriate);
\item how learned constraints are shared between workers (if learning is implemented); and
\item specific implementation details and abstractions.
\end{itemize}

\subsection{Parallel Backtracking Search}\label{sub:parbacktracksearch}

Parallel search in simple branch-and-bound settings has a long history. For example, Karp and Zhang \citeyear{karp-zhang-randpar93} proposed a randomized work allocation strategy for backtrack and branch-and-bound search. Bader et al \citeyear{BaderHC05} and Crainic et al
\citeyear{CrainicLR06} review early works, and discuss implementation issues.

First we define the key concept of a \textit{semantic path}. A semantic path is a sequence of search decisions (typically of the form \(x=a\) or \(x\ne a\) for some decision variable \(x\) and constant \(a\)). A search node of a DFS tree is uniquely (and very compactly) described by a semantic path containing all the search decisions between the node and the root. 

Perron was one of the first to report on parallel search in a commercial constraint programming toolkit
\cite{perron1999}, the 1999 update of ILOG Solver. The search space is represented as a tree of search
nodes partitioned into an explored part, a frontier and an unexplored part. When a worker enters a new node, it does so by \textit{recomputation}: each decision in a semantic path is applied to reach the starting point, propagation is performed then search begins. 
The recomputation scheme is general enough to allow more
exotic search algorithms than DFS, and it also allows nodes to be allocated to different processors on a shared memory
machine. Each processor runs its own search process exploring different parts of this common search tree, with a
communications process ensuring work balancing and termination detection. Empirical evaluation was on a 4 processor
machine using jobshop scheduling problems, i.e.\ optimisation problems rather than searching for the first solution. When using
complete search (in particular variants of LDS) parallelism showed a linear speed-up, but with depth first search the improvements
were less convincing. On the whole, the results are less than conclusive.

A similar approach was taken by Schulte \citeyear{schulte2000}, who implemented  parallel search in the multiparadigm programming system Oz/Mozart (which supports both concurrency and constraint programming). The goal was to make use of commodity computers on a network, therefore communication between machines is limited. 
As in Perron \citeyear{perron1999}, a search node is given to a worker which then explores the search tree beneath the node. Distributing search nodes is a natural choice because they can be represented very compactly as semantic paths.  The approach is work-sharing, mediated by a single manager, and with a coarse granularity. 
Results show close to linear  speedup (between a 4\% and 52\% overhead associated with distributing work), although a maximum of six workers were used. Nowadays a single machine might have many more than six cores. 

Zoeteweij and Arbab \citeyear{DBLP:conf/coordination/ZoeteweijA04} describe a component-based approach to parallel search. They design a system which requires only that solvers may publish their search frontiers, and use added constraints to direct each solver component to different areas of the search tree. Results are presented on three benchmark instances, showing speedups of between ten and fifteen on sixteen cores.

The COMET constraint programming toolkit was enhanced to allow multicore parallel search using shared memory~\cite{cp2007}. This is done
``under the hood'' so that a constraint programmer need not know that it is taking place or how it is implemented. Each
processor core runs a worker. When a worker expands its current search node, it produces new unexpanded search nodes, where an unexpanded node is a self-contained subproblem specified as a semantic path. Parallel COMET uses
a technique called \textit{work stealing} where workers who have run out of work take unexpanded nodes from other workers,
leaving them less work to do and keeping all workers busy. It is implemented as follows: Search nodes are
represented as continuations \cite{continuations} and are added to a central pool. When workers are idle they steal
continuations from the central pool, and this is synchronized; in the case of optimization problems workers communicate
new bounds on the objective function, again synchronized.  Experiments were performed on N-Queens, Scene Allocation,
Graph Colouring and the Golomb Ruler problem using depth first search and limited discrepancy search (LDS) on 1 to 4 processors.
Speed ups were a bit less than linear, although superlinear speedups occurred with LDS and this was attributed to the
order that continuations were stolen, disrupting the normal search order.

Dal Pal\`u et al~\citeyear{dalpalu2015gpu} parallelised the search process of a SAT solver using many threads on a GPU. In the context of a DPLL search without clause learning,  the host CPU explores the upper part of the search tree and distributes subtrees to the GPU cores.  They report a speedup of 38 times compared to the equivalent sequential process on the host CPU. Parallelising a SAT solver with conflict-directed clause learning is left to future work. 

\subsection{Parallel Local Search}\label{sec:parlocalsearch}

Michel, See and Van Hentenryck \citeyear{cp2006,comet-local-09} presented parallel local search in the COMET system. The papers describe an architecture for distributing the work over multiple (heterogeneous)
machines.  The distribution of tasks is intended to be nearly independent of the COMET program that describes the search.
The COMET programs given as examples describe various local search strategies including evolutionary algorithms and varieties of
constraint-based local search \cite{cblsbook}.  Parallelism is implemented by distributing different starting assignments to the workers, i.e.\ the `multi-walk' strategy of \cite{Verhoeven1995}.  The experimental results demonstrate close to linear speedups with up to 12 workers.

An extensive body of work investigates parallelising the ``Adaptive Search" \cite{diaz2001yet} local search method for constraint solving. A multi-walk approach on the magic squares, perfect squares, and all-interval series problems shows increasing speedups with more cores can be obtained, e.g. more than 100 times speedup with 256 cores \cite{Caniou2011}.  Speedups do flatten with the number of cores, and it seems that the smaller the benchmark, the faster the flattening occurs \cite{Caniou2011}. 
Even better results are obtained on the Costas Array Problem. 
In a two-dimensional Costas Array, cells must be filled in a square grid, such that there is exactly one filled cell in each row and column, and no two vectors between two filled cells are the same \cite{drakakis2006}.   
A pure multi-walk approach gave almost linear improvement in search up to as many as 8,192 cores \cite{Caniou2015}. 
This shows that foregoing the ability to share information between search processes can lead to almost perfect speedups.  
Parallelising Adaptive Search has been explored on different multi-core architectures.   We mentioned in \Cref{puils} work on using Adaptive Search using GPUs \cite{arbelaez2014gpu}.
The Cell Broadband Engine (used in the Sony Playstation 3) contains one controlling cores and a number of subsidiary ones called ``Synergistic Programming Elements" (SPEs). Adaptive Search was parallelised effectively in this context on some benchmark problems, although scalability might be limited by the small local stores on the SPEs.  
Using the Partitioned
Global Address Space programming language X10, a multi-walk approach led to good results with increasing numbers of cores, again flattening off except in the case of Costas Arrays \cite{Munera2014}.  Munera,  Diaz, Abreu, \&  Codognet \citeyear{Munera2014b}  improved results further by introducing cooperation between the separate processes, and this was successfully used to solve hard instances of stable marriage problems with ties and incomplete lists \cite{munera2015solving}.

In the context of parallel local search for SAT, Arbelaez and Hamadi \citeyear{arbelaez-hamadi-11} proposed an improvement to the standard restart strategy employed to diversify search. Each thread uses a distinct local search algorithm and all threads employ restarts.  Threads share the best assignment they have found so far (with the cost, i.e.\ the number of unsatisfied clauses). When restarting a thread, rather than restarting at a random assignment, the best assignments of other threads are synthesised into a single assigment.  They show this cooperation technique improves performance in experiments with four and eight threads. 
Arbelaez and Codognet \citeyear{arbelaez-codognet-12} studied the same cooperation strategy with higher degrees of parallelism.  They scaled their approach up to 256 threads using independent groups of up to 16 cooperating solvers. There is no communication between groups, and in this case each thread ran the same local search algorithm. The approach is evaluated on random SAT instances from a recent SAT competition, and the reported speedups are sub-linear. For example, with 256 threads where groups of four threads cooperate, the paper reports a speedup of 5.95 times compared to a single thread.  

A later publication by Arbelaez and Codognet \citeyear{arbelaez-codognet-13} contains a much broader experiment with both structured and random SAT instances drawn from a recent SAT competition. In this case a pure multi-walk approach is used, i.e.\ there is no cooperation between threads. In common with \cite{arbelaez-codognet-12} speedups on random instances are poor. However, for some structured instances they report near-linear and even super-linear speedups with 512 threads when compared to 16 threads. 

A side-benefit of using the multi-walk approach for parallel local search is that it can increase the robustness of timing of local search: with multiple independent walks, the variability in the time to find a first solution is greatly reduced  \cite{CPE:CPE1855}.  

\subsection{Search Order and Heuristics}

The issue of parallelism disrupting the search order is well-known in conventional branch and bound search, where non-linear speedups are called \emph{anomalies}:
Lai and Sahni \citeyear{DBLP:journals/cacm/LaiS84} and Li and Wah \citeyear{DBLP:journals/tc/LiW86} explain the phenomenon, and
also show how to guarantee that absolute slowdowns will never occur in a synchronous setting. This is extended to an
asynchronous setting, which more closely resembles common modern hardware, by de Bruin et al
\citeyear{DBLP:conf/irregular/BruinKT95}.  Lai and Sahni suggest that ``anomalous behaviour will be rarely witnessed in
practice'', however this claim relies on a set of assumptions that are often broken by modern CP solvers.

Chu, Schulte and Stuckey \citeyear{cp2009} refine the work stealing approach, introducing \textit{confidence-based} work stealing. The authors point out
that work stealing not only allows load balancing, but also influences the order in which the search tree is
explored. In earlier work stealing approaches (such as \cite{cp2007}) the strategy was usually to steal from as close to the root as possible. The
paper includes examples showing how it can sometimes be far better to steal ``left and low'' (i.e. as deep as
possible) but sometimes stealing high can be better. The message of course is that it should be dynamic. They claim that
the presence of a branching heuristic complicates the process of finding a good stealing strategy.  Consequently the
proposal is to steal based on confidence: the estimated distribution of solution densities among the children of a node.
When doing binary branching, this is equivalent to confidence in the heuristic (since it chooses that left branch).
Ideally the user should provide a confidence function for the heuristic, but the authors show how a simple substitute
can still work when this is not available. Confidence values are updated during search. However, in practice they found
that stealing too low tends to increase the communication cost.  Hence they set a bound above the average fail depth
below which search nodes cannot be stolen. Experimental results demonstrate the effectiveness of the technique, ranging from
speedups of 7 times to superlinear on the benchmarks in the paper.

The same observation is brought back to a conventional branch
and bound setting by McCreesh and Prosser \citeyear{DBLP:journals/topc/McCreeshP15}, where constraint programming
terminology is used to explain observed behaviour in several different parallel maximum clique algorithms. Different
work distribution strategies are compared, with the results showing that the interaction between search order and work
splitting often has more of an effect on the results than work balance. Speedups are usually at least close to linear on
64 cores, with super-linear speedups being common.

Xie and Davenport \citeyear{DBLP:conf/cpaior/XieD10} describe an attempt to
exploit an IBM Blue Gene supercomputer, which has large numbers of relatively
slow processors. They use a form of parallel limited discrepancy search, where
a master worker allocates work to other processes initially, and where workers
generate additional parallelism when they believe they are exploring a large
subtree. The exact interaction with the discrepancy ordering is not described.
Results on resource-constraint project scheduling benchmarks suggest roughly
linear scalability up to 256 processors, and sometimes up to 1,024 processors,
but Xie and Davenport believe that multiple masters would be required for
larger processor counts.

Distributed parallel discrepancy search is investigated further by Moisan et al
\citeyear{discrepancy-cp13,moisan2014parallel}, who investigate a problem involving integrated planning and scheduling for
the forest-products industry. Each worker is independent, recomputing only the subset of the search tree containing the
leaf nodes allocated to that worker. This would usually be a substantial overhead, but in their setting, leaf nodes of
the search tree are much more expensive than internal nodes, since a full linear program is solved at each leaf. Search
is not run until completion, but the solutions obtained using 4,096 workers by either form of discrepancy search are of
much higher quality than conventional parallel backtracking search, demonstrating the importance of controlling the
interaction between parallelism and value-ordering heuristics in this setting.

\subsection{Heuristic-Ignorant Decompositions}

A prototype for distributed depth-first search based upon splitting and
work-stealing is described by Jaffar et al
\citeyear{DBLP:conf/ictai/JaffarSYZ04}. To reduce contention, they use a
heuristic to attempt to steal the largest remaining job. Results are presented
on a selection of binary integer linear programming models, showing roughly
linear scalability up to 64 processors. The experiments consider only
enumeration and optimisation problems, to ``avoid any effects such as
superlinear speedup''; the importance of bounds-sharing for optimisation
problems is noted.

Kotthoff and Moore \citeyear{kotthoffMoore} describe an approach to distributed search with no direct communication between workers.
The approach assumes a job queueing system is available as a substrate. A standard solver is wrapped
in a script that does the following: 
\begin{enumerate}
\item Search until a time limit is reached or completion.
\begin{enumerate}
\item If solution found then terminate all jobs and return solution.
\item If search space completed then halt.
\item Otherwise, split the domain of the current variable (i.e.\ the variable about to be branched when the solver timed out) into \(n\) parts (where \(n\) is a constant).
\begin{enumerate}
\item To each slice add a set of restart nogoods that rules out the search space explored so far in the current solver. 
\item Submit each slice to the job server. 
\end{enumerate}
\end{enumerate}
\end{enumerate}
Hence the technique is similar to work stealing but
less dynamic---work cannot be stolen until a time limit is reached.
The novelty in this approach is twofold: using a job server to avoid implementing distribution, fault tolerance, etc.; and
using restart nogoods rather than recomputation to rule out previously explored search.
The former should be self-explanatory. The latter may not be:  recomputation implicitly assumes that the parent will continue to 
search and it is giving away only a part of its search space. 
This technique works instead by the parent giving up all its search space and splitting itself into $n$ parts. Restart nogoods 
ensure that the child processes are together given the \textit{remaining} part of the parent
search space, and made to search different portions of it by the splitting constraints.
It was not determined if this approach is better than recomputation in practice. The proposed approach is more flexible than recomputation, in the sense that the children are allowed to search the remaining space in any order, instead of always having to stick to the early part of the parent's variable ordering. This allows each solver to use
a different search strategy, but no experiments were carried out doing this.

Machado et al \citeyear{DBLP:conf/icpp/MachadoPA13} describe a distributed work-stealing approach built on top of MPI and pthreads, using a mix of local and global work pools. Results on the N-Queens enumeration problem with $n = 17$ show roughly linear speedups up to 512 processors, whilst results on a single quadratic assignment problem optimisation instance are similar. Machado et al note that their system has a non-deterministic execution time, since the amount of search depends upon when an optimal solution is found.

Constraint solvers often employ \textit{restarts} (where search is started again with a different heuristic ordering) to allow the solver to break out of large subtrees that contain no solutions. Cire et al \citeyear{cire2014parallel} noted that many constraint solvers do not learn during search, so very little or no information is retained when search is restarted, therefore the search following a restart may be performed in parallel with the search before the restart. In this way they obtained near-linear speedups with up to 32 threads. 

Fischetti, Monaci and Salvagnin \citeyear{DBLP:conf/cpaior/FischettiMS14} present a somewhat different splitting technique.  A search tree is evaluated
until sufficiently many open nodes are generated, and then workers use a deterministic rule to partition open nodes
between them. This approach requires nearly no communication between workers. An implementation in Gecode shows
speedups approaching linear using 64 cores. The authors state that since they are ``interested in measuring the
scalability of our method, we considered only instances which are either infeasible or in which we are required to find
all feasible solutions (the parallel speedup for finding a first feasible solution can be completely uncorrelated to the
number of workers, making the results hard to analyze)''. The possibility that scalability could be less important than
search order is not considered.

Malapert, R\'egin and Rezgui \citeyear{eps-jair-16} describe an approach they call Embarrassingly Parallel Search. The idea is to decompose a
problem semi-statically using a depth-bounded depth-first search, creating more subproblems than workers, and then to distribute the subproblems to the workers from a queue.
The authors show that if sufficiently many subproblems are created (30 times the number of workers is suggested), and if
it is ensured that these subproblems are not trivially detected as inconsistent, then the balance problem is addressed.
Results on optimisation and enumeration problems are presented using up to 40 cores on a single machine, and up to 512
cores across a data centre, approaching linear speedups. Results for decision problems are not presented.

Palmieri, R\'egin and Schaus \citeyear{DBLP:conf/cp/PalmieriRS16} also use the Embarrassingly Parallel Search decomposition for heuristic selection, as an alternative
to portfolios.  They perform the decomposition, use parallelism to try many variations of variable and value-ordering
heuristics on a small subset of the decomposed subproblems, and then select the most promising choice for the remainder
of the search. Results are reported on a wide range of enumeration and optimisation problems, showing that the technique
beats a multi-armed bandit portfolio.

Menouer et al \citeyear{DBLP:journals/ijpp/MenouerRCR16} show that starting with a static decomposition, and then switching to dynamic work-stealing, yield better results than either technique on its own when parallelising the OR-Tools solver. Results using two twelve core machines are mixed, with speedups ranging from seven to ten.

\subsection{Learning in Parallel Search}\label{sub:parsearchlearning}

In this section we cover approaches that parallelize the search process while also learning implied constraints that are shared among workers. The bulk of this work comes from the SAT community where Conflict-Driven Clause Learning (CDCL) SAT solvers have been very successful and dominate the field. Martins, Manquinho and Lynce~\citeyear{sat-survey-12} have written a detailed survey of approaches to parallelisation in SAT, and H\"olldobler et al \citeyear{sat-survey-icacsis11} wrote a short survey of recent developments in complete parallel SAT solvers. An earlier survey by Singer \citeyear{sat-survey-singer06} has been largely superseded by Martins et al \citeyear{sat-survey-12}. 

Martins et al \citeyear{sat-survey-12} identified multi-agent search and search-space splitting as the two main approaches to parallel SAT solving. We covered multi-agent search in \Cref{sub:multiagent-sat}, and we cover search-space splitting here. We select a small number of the most interesting contributions, and refer the reader to the earlier surveys \cite{sat-survey-12,sat-survey-icacsis11,sat-survey-singer06} for more details.

\textit{Guiding paths} are a key concept in parallel search for SAT. Guiding paths are almost identical to semantic paths (described above) in CP. The other important concept is clause sharing, where the key is to avoid sharing all learned clauses by heuristically selecting the most effective ones, as described in \Cref{sub:multiagent-sat}. 

GridSAT \cite{gridsat-chrabakh-wolski} is an early example of a parallel CDCL SAT solver. It is based on the sequential solver Chaff, and is designed to run on a heterogeneous network of computers so it makes no use of shared memory. The search is distributed using guiding paths. Each thread is able to split its portion of the search space into two pieces at any time, generating a new guiding path from the topmost unexplored search node. The approach is similar to work stealing as employed by some CP solvers.  In addition, learned clauses are shared between threads when their length is less than a limit that can be fixed or dynamically adjusted during search. Reported speedups are sub-linear for the most part, however the authors report that a number of previously unsolved SAT instances are solved by GridSAT. 

Chu and Stuckey \citeyear{pminisat} presented PMiniSat, a parallelization of MiniSat 2.0 employing work stealing. The
solver uses several techniques for sharing clauses among threads. Firstly, all clauses with length beneath a certain threshold are shared among \textit{all} threads;
this technique has been used previously, for example by ManySat \cite{manysat-09}. Secondly, a novel measure called \textit{effective length} is used: the effective length of
a learned clause is evaluated per worker, and it is the number of literals in the clause that are not false under the current assignment. A small effective length indicates that a small number of literals need to be set to false before the clause unit propagates. Only clauses whose effective length is less than a given threshold are accepted by other threads.
Hence clauses are preferentially shared between pairs of workers that are searching similar areas. Third, a worker is able to store clauses
in a suspended state until it is working on another item of work where the shared clause is unit. Unfortunately the details of this are very sketchy and no
implementation is described.
The similarities between PMiniSat and multi-agent search for SAT are clear. Finally, Hamadi et al \citeyear{DBLP:journals/jsat/HamadiJPS11} showcase an interesting variation on this theme: by careful use of barrier synchronisation, it is possible to preserve some of the deterministic aspects of sequential search in a parallel setting, even when sharing learned clauses.

Schubert, Lewis and Becker \citeyear{pamiraxt} presented PaMiraXT, a hybrid multicore and multicomputer SAT solver. On each computer it 
uses all available cores. The system solves a single SAT instance using guiding paths to differentiate the search on each worker. The workers are guaranteed to search disjoint parts of the search space. Work stealing is employed when a worker completes its search.  Once again, a key concern is how to
share clauses efficiently and effectively. On each computer, the solvers all share a single clause database and they all 
propagate all the clauses of the other processes. 
The clauses are all stored in a shared
read only memory segment and the workers keep their watches in their own memory space to avoid contention. 
To share work within one computer, there is a master process that steals work and passes it to idle workers. 
Work stealing is performed as close to the root as possible.
Between computers, the workers share clauses of length 3 and under.
The work stealing between computers is elegant. The same mechanism is used 
as within a computer, i.e.\ a 
process requests some work
from the master, but instead of solving it directly it sends the work to another computer.
Experiments clearly demonstrate that larger numbers of cores reduce the wallclock time, however the total time spent by all cores is not provided and so it is unknown whether PaMiraXT can achieve a linear or super-linear speed-up. 

Cube and Conquer~\cite{heule2011cube} is a complementary approach to search space splitting. The search space is divided into thousands or possibly millions of fragments (named \textit{cubes}) using strong lookahead heuristics to guide the choice of variables to assign. The cubes are divided among the workers and each solved independently using a CDCL algorithm. Thus Cube and Conquer is a hybrid of strong lookahead and CDCL. 
The Cube and Conquer strategy is used by Treengeling \cite{plingeling2013}, one of two parallel versions of the highly successful SAT solver Lingeling.  Treengeling came first in the parallel track of the SAT 2016 competition. 

The Cube and Conquer approach has been successfully applied to a very challenging problem in mathematics named the Boolean Pythagorean Triples Problem~\cite{heule2016pythag,heule2017pythag}. A Pythagorean triple is a set of three natural numbers \(\{a,b,c\}\) such that \(a^2+b^2=c^2\). The problem is to partition a set of natural numbers \(\{1\ldots n\}\) into two parts such that neither part contains a Pythagorean triple. For a given value of \(n\) the problem encodes straightforwardly into SAT, and Cube and Conquer was applied to prove that \(\{1\ldots 7824\}\) can be partitioned whereas \(\{1\ldots 7825\}\) cannot. Solving both instances (\(n=7824\) and \(n=7825\)) took 35,000 hours of CPU time on a cluster with 800 cores, enabling the instances to be solved in about 2 days. 

Ehlers and Stuckey \citeyear{DBLP:conf/cpaior/EhlersS16} parallelized a \textit{lazy clause generation} constraint programming solver. Lazy clause generation solvers employ SAT-style learning with CP propagation algorithms.  They demonstrate that a hybrid approach (combining SAT-style multi-agent search with CP-style
parallel search) provides the best results, although it does not scale as smoothly as classical CP parallel search. When
finding an optimal solution, speedups are often super-linear, whereas when proving optimality or unsatisfiability,
speedups are usually sub-linear.

Dovier et al \citeyear{DBLP:conf/padl/DovierFPV16} implemented an entire ASP solver on a GPU, using nVidia's CUDA framework. The results are extremely mixed: the sequential CLASP solver on a CPU vastly outperforms the GPU solver. However, CLASP with heuristics disabled is occasionally comprehensively beaten by the GPU solver (which does not implement heuristics). It is not clear whether these successes are due to search ordering coincidences that heuristics would address, or whether the GPU solver would be genuinely competitive if it implemented heuristics itself. Nonetheless, these results demonstrate that it is at least \emph{possible} to run complex search algorithms on a GPU, even if it is not clear whether doing so will ever be beneficial.

Finally we refer to Hamadi and Wintersteiger's seven challenges in parallel SAT \citeyear{hamadi2013seven}. The second challenge is the most relevant here: to develop improved dynamic decompositions, either splitting the search space or the instance. Cube and Conquer employs static decomposition, however its success on very difficult instances of the Boolean Pythagorean Triples Problem (one satisfiable, one unsatisfiable), and its first place in the SAT competition suggests that it addresses the challenge well. 

\subsection{Richer Search Trees}

An early work on parallel search is due to Bill Clocksin's DelPhi principle \cite{bill}.  The context here
is Prolog, which explores AND-OR trees, which are richer than the OR-trees common in constraint
solving. The motivation is to avoid the overhead incurred by having a shared memory or copying
computation state between processors. The central idea is to associate a processor with each {\em path} through the
search tree, hence avoiding overhead due to transferring state between processors mid-path. However this seems like a
false economy as it results in duplicated work (consider two branches that differ only at the very bottom).  Usually we
don't have enough processors to allocate one to each of the possible paths. Suppose we have $n$ processors and we explore all
log$_2n$ paths. If we find a solution, terminate otherwise consider the log$_2k$-bit extensions to these paths (where
$k$ is the number of processors we have left---there is some art and strategy to this as $k$ will continue to increase
as the original log$_2n$ bit paths are explored) and continue. In fact to continue DelPhi search starts from scratch
again in order to be in the right state (the paths are stored very compactly as bit strings).

Much more recently, Bergman et al \citeyear{bergman2014parallel} worked on decision diagram search trees. When decision diagrams
become too wide, usually a number of subproblems are created and evaluated sequentially. Instead, in this work they are evaluated in parallel. Nearly linear speedups are obtained on maximum independent set instances using up to 256 cores, but it is
worth noting that not a single one of these results from 256 cores beats a better sequential algorithm with one core
\cite{DBLP:journals/topc/McCreeshP15}.

Finally, Otten and Dechter \citeyear{DBLP:journals/jair/OttenD17} describe a parallelisation of AND-OR search trees for
genetic linkage applications across a grid computing system based around the CondorHT framework. This system does not
allow dynamic communication between nodes, and so considerable effort is spent to obtain a good decomposition upfront.
Otten and Dechter demonstrate a machine learning approach which is able to predict a suitable partitioning depth. They
also discuss redundancies caused by work duplication, which are necessary due to the inability to share information
during search. Their experiments scale to a cluster of 27 machines, each with 12 cores (for a total of 324 cores);
speedups vary considerably, but are often comfortably over 100.

\section{Algorithm Portfolios}\label{sec:portfolios}

Algorithm portfolios were originally conceived \cite{huberman1997economics} as a means to exploit and guard against the variability in performance found in combinatorial search algorithms with a stochastic component, for example in the heuristic selection of variables and values to guide the search. Huberman et al. \citeyear{huberman1997economics} characterise the `reward' of such an algorithm as its mean performance, while its `risk' is the variance in that performance. Taking inspiration from Economics, they present a method for allocating computational resources across a portfolio of algorithms each independently solving the same problem instance so as to balance risk and reward on an `efficient frontier'. Gomes and Selman \citeyear{gomes2001algorithm} follow a similar approach to consider how to assign algorithm instances (including multiple instances of the same algorithm) across a number of processing units, demonstrating empirically that the ideal portfolio composition varies according to the number of processing units available.

The simplest method of parallel portfolio construction is the hand-selection of a set of solvers, combined with a scheme for how these solvers should distributed among the available computing resources. This approach is taken by {\em ppfolio} \cite{RousselPPFolio}, which employs five SAT solvers, originally on the basis of their performance of the 2009 SAT solving competition and then subsequently on more recent competitions, and specifies the combination of these solvers to run for a given number of available cores. A similar approach is taken by {\em pfolioUZK} \cite{wotzlaw2012pfoliouzk}, which combines both complete and incomplete SAT solvers. {\em ppfolio}-like portfolios are termed Plain Parallel Portfolios by Aigner et al. \citeyear{aigner2013analysis}, who perform an empirical analysis of their performance and scalability. A theoretical analysis of the success of the plain parallel approach is given by Hyv\"{a}rinen and Manthey \citeyear{hyvarinen2012designing}. Inspired by {\em ppfolio}, {\em aspeed} \cite{hoos2015aspeed} automates the construction of  portfolios from benchmark data for a particular class of problems. The task of allocating time slices on processing units to solvers so as to minimise timeouts with respect to the benchmark data is formulated and solved as an answer set programming problem, enabling the automated production of portfolios with non-uniform resource allocation.

An alternative method of portfolio construction is on a per-instance basis, depending upon the nature of the instance to be solved. In this approach, problem features thought to be useful performance predictors are selected and a performance model is built based upon these features and a set of training problem instances. The model is then used to make selection and resource allocation decisions based on the features of an unseen instance. This method can be viewed as an example of the Algorithm Selection Problem \cite{rice1976algorithm}.

In the sequential context, a prominent early example of this approach is SATzilla \cite{xu2008satzilla}, which learns an empirical hardness model that predicts the runtime for an algorithm on an instance based on the instance's features. {\em p3S} \cite{malitsky2012parallel}, a parallel version of the earlier sequential {\em 3s} \cite{kadioglu2011algorithm}, employs the same set of features as SATzilla. Measured via Euclidean distance in the normalised feature space, {\em p3S} uses the $k$ nearest neighbours in the training set to the instance to be solved to decide the composition of a parallel portfolio, which may contain both sequential and parallel SAT solvers. Also following SATzilla, this portfolio is supplemented by a statically decided parallel solver schedule designed to consume the first ten percent of available run-time in an attempt to avoid expensive per-instance feature computation. Both decisions are formulated and solved via Integer Programming. CSHC \cite{malitsky2013algorithm} employs the same static schedule as {\em 3S} for the first ten percent of available run time, followed by an instance-based selection made by matching the incoming instance against statically formed clusters of training instances. In this case, clustering is performed in a hierarchical manner, incorporating cost information. A parallel version of CSHC ran in the 2013 SAT competition \cite{malitsky2013parallel} deploying three separate instances of CSHC, each trained on a different category of the competition instances. Lindauer et al. \citeyear{lindauer2015sequential} demonstrate how to extend existing sequential algorithm selectors to rank candidates so as to construct a parallel portfolio from the top-ranked solvers. They also use {\em aspeed} to compute parallel pre-solving schedules.

In Constraint Programming, CPHydra \cite{o2008using} employs case-based reasoning, storing feature-annotated instances and performance data in a case base, then retrieving the most similar case to an unseen problem instance. Yun and Epstein \citeyear{yun2012learning} extend CPHydra to work in parallel, employing heuristics to create parallel portfolios from single-processor portfolios. Similarly to {\em 3s}, {\em sunny-cp} \cite{amadini2015sunny} employs a $k$ nearest neighbour approach to select relevant training instances, based upon which solvers are selected, allocated a certain runtime and scheduled. {\em sunny-cp2} \cite{amadini2015multicore} generates a parallel portfolio on $c$ cores from a {\em sunny-cp} schedule by allocating to the first $c-1$ cores the solvers scheduled with the greatest allocated time and allocating the remainder to the final core so as to preserve their original schedule. {\em sunny-cp2} can also solve constrained optimisation problems by supporting bounds communication between solvers.

Bordeaux et al. \citeyear{ijcai2009} highlight the importance of sources of variability when the number of processing units is large. They identify three desirable qualities: scalable (different settings will result in different runtimes); favourable (varying from the default does not systematically worsen performance); solver-independent (exploiting the features of individual solvers prohibits re-usability). One such source of variability is in the configuration space of modern constraint or SAT solvers. The ACPP system \cite{lindauer2017automatic} exploits this opportunity to construct parallel portfolios automatically from configurations of individual sequential solvers, sets of sequential solvers, or a combination of sequential and parallel solvers. Proteus \cite{hurley2014proteus} provides further variety in the constituents of a portfolio by considering different encodings of a given problem (in this case into SAT) and different potential solvers for each. Similarly, Akg\"un et al. \citeyear{akgun2010refining} discuss the automated refinement via {\sc Conjure} \cite{akgun2011extensible,akgun2014extensible,wetter2015automatically}, which captures patterns in constraint modelling such as symmetry breaking \cite{frisch2007symmetry,akgun2013automated,akgun2014breaking} and matrix modelling \cite{flener2001symmetry,flener2002matrix}, of multiple constraint models from an abstract specification in the {\sc Essence} \cite{frisch2005essence,frisch2007design,frisch2008ssence} language in order to form constraint model portfolios. \textsc{Savile Row} \cite{sr-journal-17} prepares the generic constraint models produced by \textsc{Conjure} for particular constraint solvers as well as output to SAT. In doing so it applies a number of optional reformulations, such as common subexpression elimination \cite{gent2008common,nightingale2014automatically,nightingale2015automatically} and the addition of implied constraints \cite{colton2001constraint,frisch2004symmetry,charnley2006automatic}, which could serve as an additional source of diversity in portfolio construction.

For further reading on algorithm selection and algorithm portfolios, see extensive surveys by Kotthoff \citeyear{kotthoff2014algorithm} and
Smith-Miles \citeyear{smith2009cross}.

\section{Conclusion}

We have presented a survey of the literature on parallel constraint solving. There is a great variety of interesting work in this area, and we divided it into four broad categories: parallel consistency and propagation, parallelizing the search process, multi-agent search, and portfolios. In each area there are challenges, such as Kasif's \citeyear{kasif90} proof of the P-completeness of establishing arc-consistency. Challenges remain in parallel SAT solving also \cite{hamadi2013seven}. However, recent results with work stealing, partitioning of the search space, portfolio approaches, and embarrassingly parallel search, give considerable cause for optimism. 

Multicore computing is now the norm. Whether we would prefer the comfort of a faster single-processor world or not, therefore, we must embrace this paradigm. There seems to be little overall guidance that can be given on how best to exploit multicore computers to speed up constraint solving. We hope at least that this survey will provide useful pointers to future researchers wishing to correct this situation.

\section{Acknowledgements}

We thank Christopher Jefferson for a number of useful discussions. We thank reviewers of this journal for their suggestions which greatly improved this paper. Finally, we thank  Enrico Pontelli for kindly granting an extension to the deadline when multiple  authors of this paper were (in parallel) suffering illness or other problems.  

\bibliographystyle{acmtrans}
\bibliography{references}

\end{document}